\documentclass[pdflatex,sn-mathphys-num]{sn-jnl}

\usepackage{graphicx}%
\usepackage{multirow}%
\usepackage{amsmath,amssymb,amsfonts}%
\usepackage{amsthm}%
\usepackage[title]{appendix}%
\usepackage{xcolor}%
\usepackage{textcomp}%
\usepackage{manyfoot}%
\usepackage{booktabs}%
\usepackage{algorithm}%
\usepackage{algorithmicx}%
\usepackage{algpseudocode}%
\usepackage{tabularx}%
\usepackage{adjustbox}%
\usepackage{listings}%
\usepackage[acronym]{glossaries}%

\theoremstyle{thmstyleone}%
\theoremstyle{thmstyletwo}%

\theoremstyle{thmstylethree}%

\begin{document}

\title[Article Title]{From research to clinic: Accelerating the translation of clinical decision support systems by making synthetic data interoperable}

\author*[1]{\fnm{Pavitra} \sur{Chauhan}}\email{pavitra.chauhan@uit.no}
\author[2]{\fnm{Mohsen} \sur{Gamal Saad Askar}}
\author[2]{\fnm{Kristian} \sur{Svendsen}}
\author[3]{\fnm{Bj{\o}rn} \sur{Fjukstad}}
\author[4]{\fnm{Brita} \sur{Elvev\r{a}g}}
\author[1]{\fnm{Lars Ailo} \sur{Bongo}}
\author[1]{\fnm{Edvard} \sur{Pedersen}}

\affil[1]{Department of Computer Science, UiT The Arctic University of Norway, Troms{\o}, 9019, Norway}
\affil[2]{Department of Pharmacy, UiT The Arctic University of Norway, Troms{\o}, 9019, Norway}
\affil[3]{DIPS AS, Troms{\o}, 9010, Norway}
\affil[4]{Department of Clinical Medicine, UiT The Arctic University of Norway, Troms{\o}, 9019, Norway}


\abstract{
\textbf{Background:} The translation of clinical decision support system (CDSS) tools from research settings into the clinic is often non-existent, partly because the focus tends to be on training machine learning models rather than tool development using the model for inference. To develop a CDSS tool that can be deployed in the clinical workflow, there is a need to integrate, validate, and test the tool on the Electronic Health Record (EHR) systems that store and manage patient data. Not surprisingly, it is rarely possible for researchers to get the necessary access to an EHR system due to legal restrictions pertaining to the protection of data privacy in patient records. We propose an architecture for using synthetic data in EHR systems to make CDSS tool development and testing much easier.

\textbf{Methods:} In this study, the architecture is implemented in the SyntHIR system. SyntHIR has three noteworthy architectural features enabling (i) integration with synthetic data generators, (ii) data interoperability, and (iii) tool transportability.

\textbf{Results:} The translational value of this approach was evaluated through two primary steps. First, a working proof-of-concept of a machine learning-based CDSS tool was developed using data from patient registries in Norway. Second, the transportability of this CDSS tool was demonstrated by successfully deploying it in Norway’s largest EHR system vendor (DIPS).

\textbf{Conclusion:} These findings showcase the value of the SyntHIR architecture as a useful reference model to accelerate the translation of “bench to bedside” research of CDSS tools. }


\maketitle

\section{Background}\label{sec:background}

\begin{figure*}[t]
\centering
\includegraphics[width=\textwidth, height=225mm, keepaspectratio]{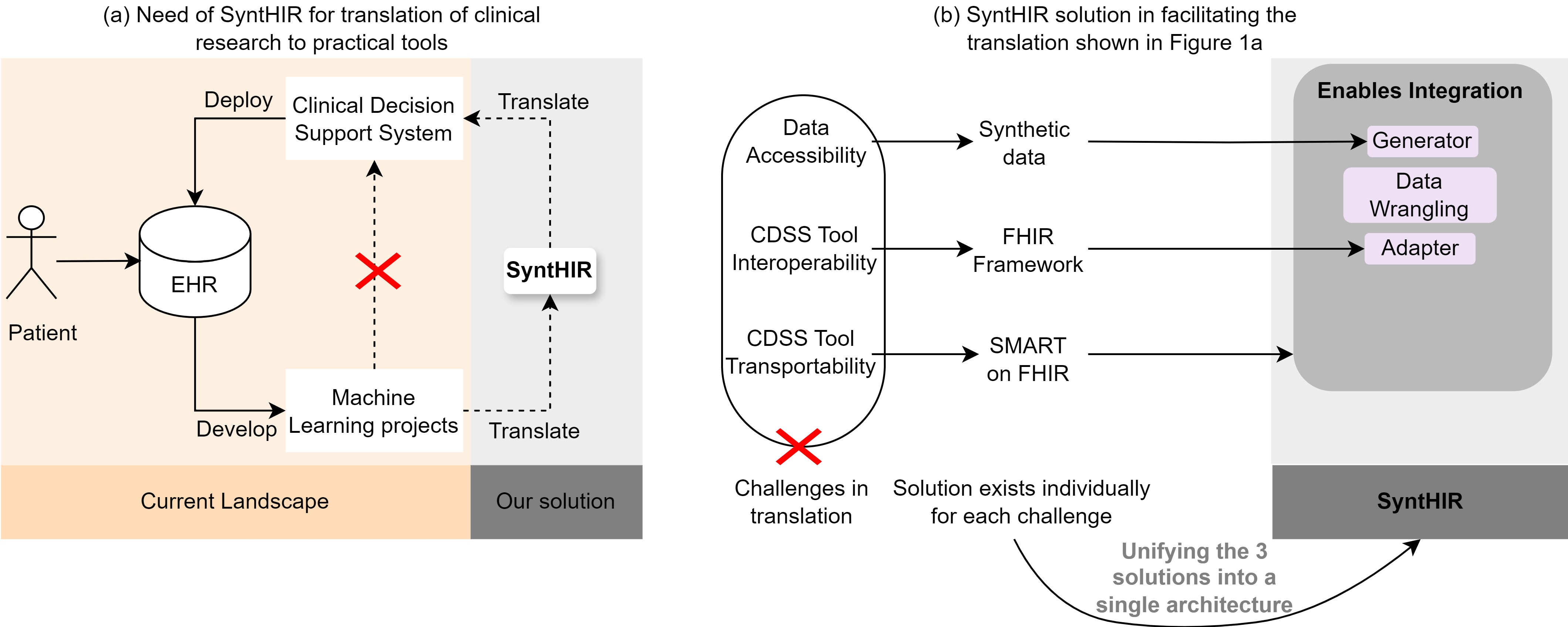}
\caption{SyntHIR system for translating CDSS tools from research to clinic. \textbf{(a)} In the current landscape, numerous machine learning projects utilize EHR data to develop novel models. These models need to be translated into a decision support system that can be deployed into an EHR system to aid physicians with decision-making. However, translating these models into practical tools presents various challenges. SyntHIR provides an architecture that facilitates the translation of machine learning models into practical CDSS tools. \textbf{(b)} The synthetic data, FHIR framework, and SMART on FHIR address the three primary challenges of translating models into tools separately, highlighting the need to integrate these into a single architecture, which is achieved via SyntHIR.
}
\label{fig:synthir_system}
\end{figure*}

Significant advancements in the research and development of machine learning models have been made through the digitization of health data \cite{jiang2011study, bashiri2024development}. However, these developments still need to be translated into clinical decision support system (CDSS) tools to improve patient healthcare outcomes \cite{zhang2022shifting}. While CDSS tools have the potential to aid physicians in decision-making \cite{susanto2023effects}, research typically focuses on developing novel methods and models \cite{ramgopal2023artificial, van2024ai, choi2023development, shaikh2021artificial} rather than translating them into realistic CDSS tools. Some machine learning-based CDSS tools have demonstrated their utility in mortality prediction and diagnosis of COVID-19 \cite{afrash2022machine, karthikeyan2021machine} and identifying medication-related errors \cite{corny2020machine, van2024ai}. Additionally, practical machine learning-based CDSS(s) such as Medicalis, Medi-span, and HERA-MI have already been shown to be clinically helpful in a variety of domains ranging from ordering radiology reports to enhancing operational efficiency and reducing unnecessary imaging costs and the early detection of breast cancer respectively \cite{medicaliscdss, khanna2020review}. The digitized health data has vast hitherto untapped potential for developing various solutions to assist physicians with clinical decision-making using novel machine learning-based CDSS tools. Unfortunately, the models are developed using a limited number of characterized datasets and typically focus on model performance and robustness, whereas integration testing with the healthcare infrastructure is needed for translation from bench to bedside of machine learning-based clinical tools \cite{mechelli2020models}. However, the actual testing and evaluation of these solutions in the clinical environment is typically not addressed \cite{sutton2020overview, gordon2020beyond, mathews2019digital}. As a result, machine learning models are not utilized to benefit patients in terms of providing higher quality healthcare and improved medical decisions (Figure \ref{fig:synthir_system}a). 

The three major challenges that limit the progress and eventual implementation of CDSS tools are data accessibility, CDSS tool interoperability, and CDSS tool transportability. Firstly, data protection regulations such as the General Data Protection Regulation (GDPR) and the Health Insurance Portability and Accountability Act (HIPPA) impose stringent requirements that complicate the process of clinical data sharing across organizations and institutions. Even if health data was accessible for researchers creating the machine learning models, developers often lack access to data for testing and debugging the CDSS tool. Secondly, it is necessary to understand and systematically implement the structure of health data sources required by EHR systems to translate a machine learning model into a functional CDSS tool. Finally, EHR systems are not easily accessible to researchers and developers, and it is necessary to test the CDSS tool on a different system before deploying it in the clinical context. Therefore, there is a need to develop and test the CDSS tool by implementing the model using the infrastructure of an EHR system but without data restrictions in production systems. The aforementioned challenges and their existing solutions are elaborated below.

\textbf{Data accessibility} The limited accessibility to health data has resulted in an increased interest among healthcare researchers and developers in generating and using synthetic data for developing machine learning-based clinical tools. Commercial platforms such as Mostly AI \cite{mostlyaisynth}, Syntegra \cite{mendelevitch2021fidelity}, and open-source solutions such as Gretel \cite{noruzman2021gretel}, Synthea \cite{walonoski2018synthea}, and ChatGPT \cite{chatgptsynth} are available for generating synthetic data. 

\textbf{CDSS tool interoperability} The CDSS tool needs to be integrated with a Fast Healthcare Interoperability Resources (FHIR \footnote{\url{https://www.hl7.org/fhir/}}) - based EHR system as all the major EHR systems have adopted the FHIR framework. This interoperability problem can be solved using an FHIR server during the development phase \cite{semenov2018patients, semenov2020experience, dullabh2022challenges, suraj2022smart}. Some examples of FHIR servers are Google Cloud Healthcare API \cite{googlecloudfhir} and Microsoft Azure API for FHIR \cite{micrsoftcloudfhir}. These servers facilitate the testing of CDSS tool integration before deploying them into another EHR system.

\textbf{CDSS tool transportability} The complexity of integrating CDSS tools across different EHR systems can be mitigated by using Substitutable Medical Applications and Reusable Technologies (SMART), commonly known as SMART on FHIR \cite{mandel2016smart}. SMART on FHIR provides an open and interoperable Application Programming Interface (API) to build health applications using FHIR standards. 



The synthetic data generators, the FHIR framework, and SMART on FHIR address the above discussed three primary challenges of data accessibility, CDSS tool interoperability, and CDSS tool transportability, respectively. SyntHIR builds on these foundations by integrating these solutions into a unified, novel architecture. There is a need to develop a novel architecture by combining existing solutions into a cohesive framework, providing a realistic development environment for machine learning-based CDSS tools using synthetic data, as illustrated in Figure \ref{fig:synthir_system}b.



\begin{table*}[th]
\centering
    \begin{adjustbox}{max width=\textwidth}
       \begin{tabular}{l  c  c  c  r}
           \toprule
             \textbf{SyntHIR Component} & \multicolumn{1}{c}{\textbf{API Purpose}} & \textbf{Request Type} & \textbf{Parameters} & \textbf{Request Body} \\
             \midrule
             \multirow{2}{*}{Data Wrangling}  & Convert to FHIR & POST & - & CSV file (to convert) \\
                                              & Convert to CSV &  POST & - & FHIR resources (to convert) \\
             \midrule
             \multirow{2}{*}{FHIR Adapter} & Upload  & POST & FHIR server URL & FHIR resources (to upload)   \\
                                           & Download 	& GET & FHIR server URL & - \\
             \midrule
             Synthetic FHIR Data Generator & Generate synthetic records & POST & Number of records & CSV file \\
             \bottomrule
       \end{tabular}
    \end{adjustbox}
\caption{API details of the functionalities implemented by the SyntHIR components}
\label{tab:synthir_api_details}
\end{table*}



\section{Methods}\label{sec:methods}

Developing a unified architecture further necessitates two requirements. First, a considerable amount of manual data wrangling is required to use any synthetic data within an FHIR-based EHR system. Second, interacting with the FHIR servers requires maintaining relationships between entities within the FHIR framework such that the data is internally consistent. The resultant unified architecture provides seamless integration of synthetically generated data from existing platforms into these EHR systems, aiding the process of developing, testing, and validating CDSS tools, as stated below:

\begin{itemize}
    \item A synthetic FHIR data generator that incorporates an open-source platform generating synthetic data resolves the issue of data accessibility to researchers and developers. In addition, it can also generate any missing fields in the data that are required by the CDSS tool. 
    \item A data wrangling component facilitates the translation between health data and the consistent data formats necessitated by the FHIR framework. Additionally, an FHIR adapter interfaces with a cloud-based FHIR server, transforming the data to fit the FHIR framework data model. 
    \item Using the SMART on FHIR framework to build CDSS tools ensures transportability across different EHR systems and enables integration with multiple EHR servers.   
\end{itemize}

The above novel architecture is implemented in a system called SyntHIR. The SyntHIR system bridges the gap between an FHIR server and synthetic data generation tools, enabling the development of CDSS tools without accessing sensitive data. Combined, these elements allow us to simulate a clinical EHR system outside the clinical setting while providing realistic health data for developing and testing CDSS tools.

\subsection{SyntHIR Components}
\label{subsec:synthir_components}

\begin{figure*}[t!]
\centering
\includegraphics[width=\textwidth, height=225mm, keepaspectratio]{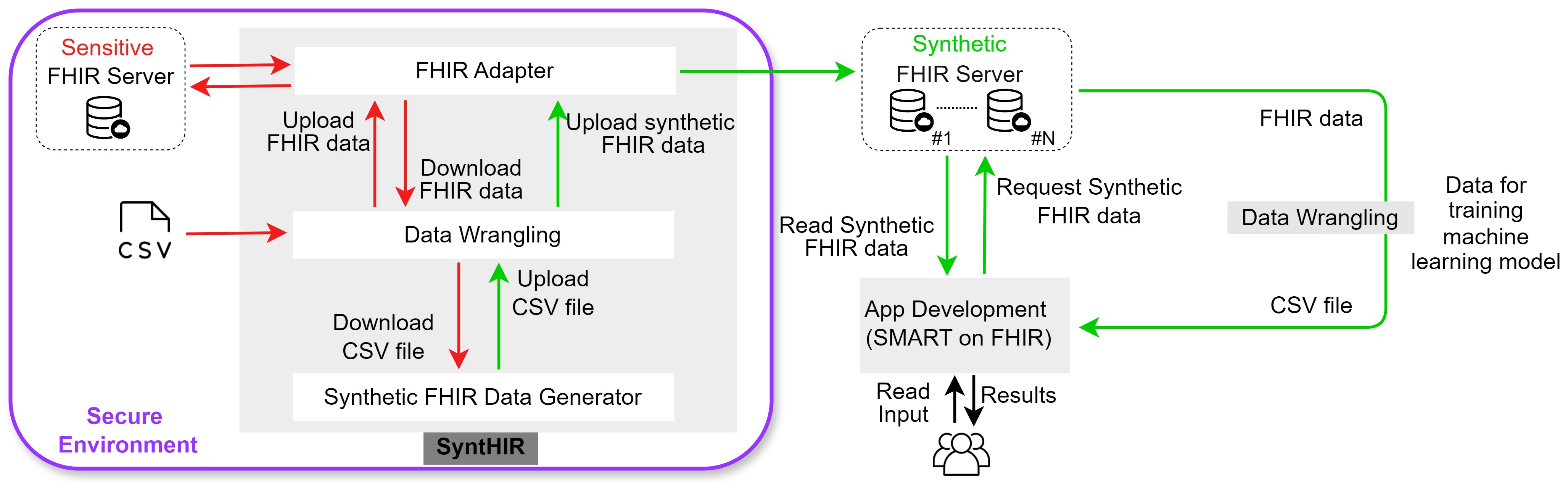}
\caption{SyntHIR architecture. Data Wrangling converts CSV health data files into FHIR format and vice versa. The  FHIR Adapter interfaces with the FHIR-based EHR servers to upload and download data from these servers. The Synthetic FHIR Data Generator generates synthetic data, and the app (SMART on FHIR) reads data from the synthetic FHIR server. The confidential data are uploaded to the sensitive FHIR server and remain in a secure environment. The arrows connecting the components depict the data flow across the components. Here, the red and green arrow denotes the flow of real and synthetic data, respectively.}
\label{fig:synthir_architecture}
\end{figure*}

The SyntHIR architecture consists of three components, namely, Data Wrangling, an FHIR Adapter, and a Synthetic FHIR Data Generator, as illustrated in Figure \ref{fig:synthir_architecture}. The design of these components is modular and provides access to their functions via APIs (see Table \ref{tab:synthir_api_details}). The restricted real patient records and synthetic datasets are stored and accessed through distinct servers, specifically, the sensitive FHIR server and the synthetic FHIR server. In our implementation, we deployed both FHIR servers on a Microsoft cloud healthcare service called `Azure API for FHIR'. All three SyntHIR components and the sensitive FHIR server are deployed in a secure environment, as shown in Figure \ref{fig:synthir_architecture}, ensuring that sensitive data is protected.



\begin{figure}[t!]
\centering
\includegraphics[width=\textwidth, height=225mm, keepaspectratio]{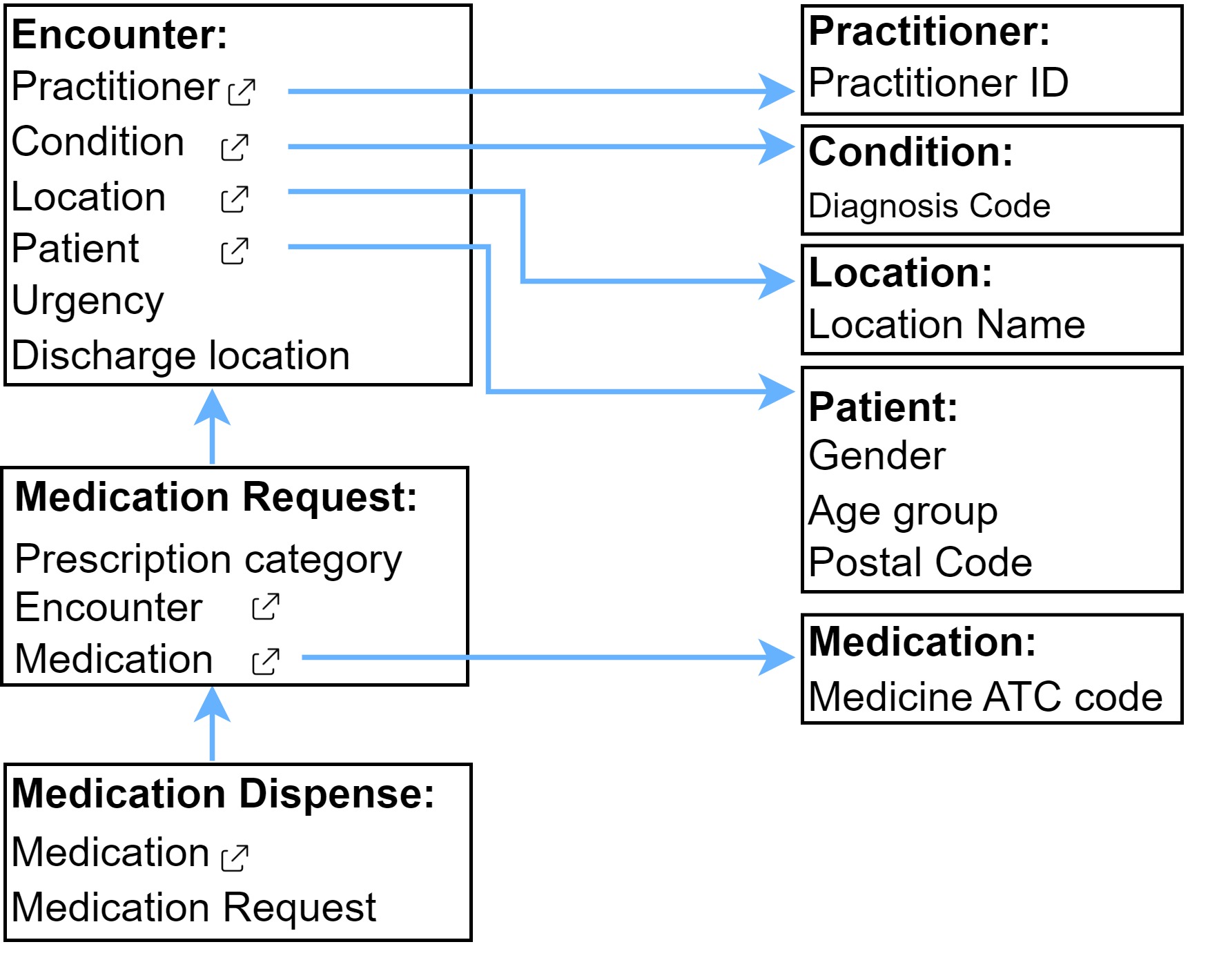}
\caption{The FHIR Resources Schema of the dataset used for the hospitalization prediction CDSS tool. Each box represents an FHIR resource with the resource name on top and the attributes listed below. The relationship between resources is depicted with lines connecting the resources. The direction of the arrow indicates the target FHIR resource retrieved from the source, and the source is represented by the link icon. The detailed schema with all the attributes of the dataset is provided in section 2 of the Supplementary.}
\label{fig:fhirschema_registry_data}
\end{figure}

\subsubsection{\textnormal{Data Wrangling:}} 
\label{subsec:data_wrangling}

The FHIR-based EHR system requires the data to be in the FHIR format. Therefore, we introduce a Data Wrangling component in the SyntHIR system. The wrangling component simplifies this process by translating the data (real or synthetic) from CSV format to FHIR data and vice versa (here, we use the terms FHIR data and FHIR resources interchangeably). Data Wrangling design consists of two parts, namely, template and mapping. The template contains the blueprint of FHIR resource contents corresponding to the CSV file. The mapping file holds the key-value pair to populate the templates from the CSV file. Here, the key is the attribute name of the CSV file, and the value is the corresponding attribute in the templates. The template and mapping files are stored in the Data Wrangling component. The template is implemented using handlebarjs \footnote{\url{https://handlebarsjs.com/}} (Supplementary section 1.1 has additional details). The Data Wrangling component is used whenever a CSV file is uploaded to the FHIR server(s) or synthetic data is generated. In addition, Data Wrangling can also be used as a standalone component to convert synthetic FHIR data into a CSV file. 

\subsubsection{\textnormal{FHIR Adapter:}} 
\label{subsec:fhir_adapter}

Uploading and downloading FHIR resources from the FHIR server(s) requires maintaining the relationship between these resources using references represented by the Uniform Resource Locators (URLs). These URLs are identifiers for the resources and are specified in the blueprint. A component called the FHIR Adapter is designed to maintain these relationships. The FHIR Adapter interfaces with both FHIR server(s) and maintains an environmental file that contains credentials to connect to the FHIR server(s). This ensures that both servers are segregated and the real data stays within the secure environment. The adapter facilitates uploading and downloading FHIR resources from Data Wrangling to FHIR server(s) and vice versa. This component accepts the URL of the FHIR server and FHIR resources to upload, or it receives the URL of the FHIR server from which it downloads all the FHIR resources. More details in section 1.2 of the Supplementary.

\subsubsection{\textnormal{Synthetic FHIR Data Generator:}} 
\label{subsec:synthethic_fhir_data_generator}

Synthetic data needs a generator platform, and it is also required to integrate the platform (with SyntHIR) within a secure environment to protect sensitive data. Therefore, the synthetic data generator platform is deployed within the SyntHIR system in a component called Synthetic FHIR Data Generator. In SyntHIR, this component interacts with Data Wrangling to provide appropriate data formats to the synthetic data generator. It receives sensitive data (CSV file) and the required number of synthetic records to generate, and outputs the synthetic data to Data Wrangling in a CSV format. Additionally, this component can impute missing values using synthetic data. We use Gretel \footnote{\url{https://gretel.ai/synthetics}}, an open-source generator platform, to generate synthetic data based on the statistical properties of the sensitive dataset (More details in section 1.3 of the Supplementary). However, any other platform can be used.

\subsection{FHIR Resource Schema}
\label{subsec:FHIR_Resource_Schema}

Data Wrangling converts each record of the input CSV to FHIR resources. Therefore, we have a list of FHIR resources, each corresponding to one record of the input CSV. The schema of FHIR resources of the dataset (NPR and NorPD) used for the hospitalization prediction CDSS tool is shown in Figure \ref{fig:fhirschema_registry_data}. The resources include Patient, which stores information about the individual receiving care; Practitioner, which holds details about the care provider objects; Location, which describes the place where healthcare services are provided; Encounter, which documents the interactions between the patient and healthcare provider; Condition, which records clinical diagnosis; Medication Request, which details the prescriptions; Medication, which identifies the prescribed medications; and Medication Dispense, which provide information about the dispensing of the medications. The attributes of each resource are also presented in Figure \ref{fig:fhirschema_registry_data}. The resources are linked through a reference URL defined as an attribute. Encounter is the central resource of the schema, which is directly related to resources such as Patient, Practitioner, Condition, and Location through its reference URL. The Medication Request resource is further directly linked to Medication and Encounter, and Medication Dispense is linked to the Medication Request resource.

\subsection{Working of CDSS tool}
\label{subsec:cdss_tool_function}

A basic CDSS tool is developed using the SMART on FHIR framework \cite{mandel2016smart} to demonstrate the functionality of SyntHIR. The tool is deployed on the cloud as a web application. It predicts the risk of hospitalization using a machine learning model. This tool can be connected to any FHIR-based EHR system (as demonstrated in section \ref{subsec:deploy_dips_arena}). Users input a patient identifier, which the tool uses to retrieve patient information, hospitalization details, prescriptions, conditions, and medications. The machine learning model uses eight input variables (more details in 3 of the Supplementary) to predict hospitalization risk. The user selects details from hospitalization, medication, and prescription options via the GUI, and the tool prompts the prediction. If the EHR data lacks any required variables, the tool retrieves them from the synthetic FHIR server. It is important to note that these missing variables do not correlate with synthetic data; they are simply imputed to ensure the tool functions correctly.

\subsection{Dataset}
\label{subsec:dataset}

The synthetic data was derived from an anonymized patient dataset consisting of Norwegian patients aged 65 and older. This real dataset was part of a project to study medication use in the elderly and its association with hospitalization. The project has ethical approval from the Regional Committees for Medical and Health Research Ethics in Norway (REK-Nord number: 2014/2182). To ensure that the synthetic version could not reasonably be identical to any real-life individuals, the dataset was anonymized prior to use in this study. The anonymization process performed by the original data custodians was done as follows. First, only relevant information was kept from the original dataset. Second, individuals' birth year and gender were replaced with randomly generated values within a plausible range of values. Finally, all dates were replaced with randomly generated dates. This resulting dataset was sufficiently scrambled to generate a synthetic dataset and was only used for the current study.



\input\begin{figure*}[t!]
\centering
\includegraphics[width=\textwidth, height=225mm, keepaspectratio]{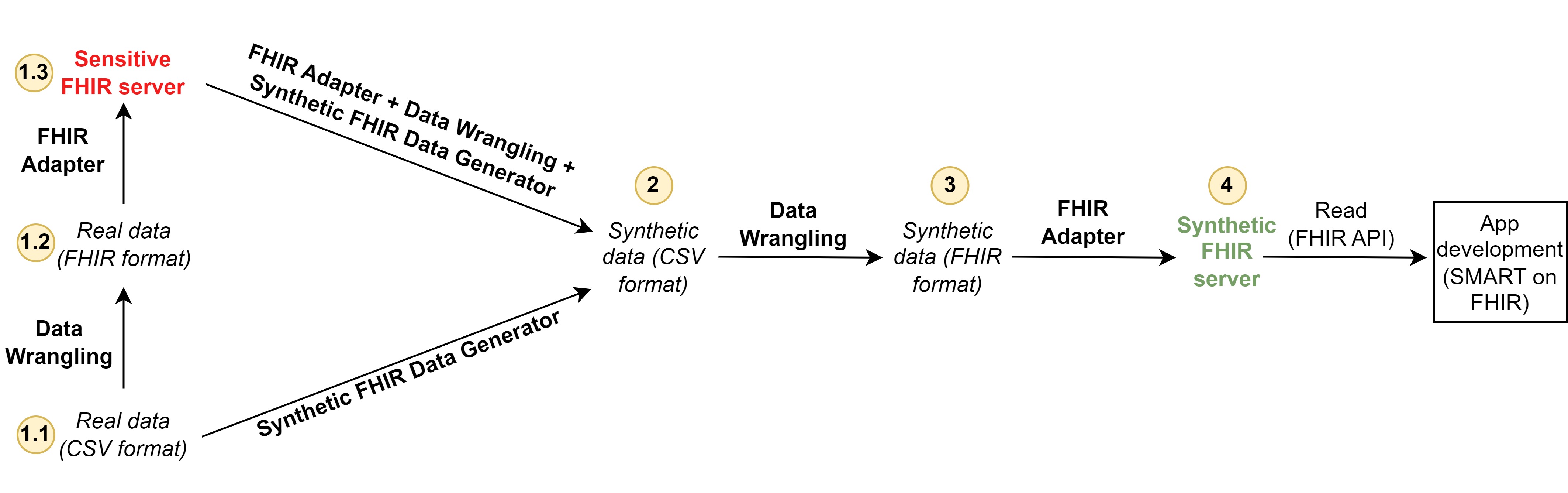}
\caption{Data Flow during CDSS tool development using SyntHIR. The arrows represent the components of the architecture used (written in bold), and the text connecting the arrows is the input and output of the components (written in italics). SyntHIR synthetic FHIR server holds the synthetic data. The FHIR API is the API provided by the FHIR server, which interfaces with the App development environment (CDSS tool). SyntHIR thus provides realistic data access for the CDSS tools. The numbering (or steps, denoted in yellow circles) indicates the different possible flow sequences of data for tool development, and the components are explained in detail in section \ref{subsec:synthir_components}. The data flow can arrive at step 2 via any of the three possibilities, i.e., 1.1, 1.2, and 1.3. }
\label{fig:synthir_app_development_flow}
\end{figure*}


\section{Results}
\label{sec:results}

The SyntHIR architecture results in three modular components, namely, Data Wrangling, an FHIR Adapter, and a Synthetic FHIR Data Generator, discussed in section \ref{subsec:synthir_components}. We validate the SyntHIR system by developing a proof-of-concept machine learning-based CDSS tool that predicts a patient's risk of hospitalization. More details about the CDSS tool's working are in section 2 of the Supplementary. We demonstrate the interoperability and transportability of the resulting CDSS tool by deploying it on the DIPS Arena \cite{dips}, the DIPS EHR system. However, our aim is not to evaluate the predictive performance and validation of the machine learning model used by the CDSS tool, but to demonstrate the tool development and testing using interoperable synthetic data.

\subsection{SyntHIR Workflow: Development of a CDSS Tool for Predicting Risk of Hospitalization}
\label{subsec:synthir_workflow}

We demonstrate the SynthHIR system by developing a CDSS tool (also termed an app in this article) using interoperable synthetic data. The CDSS tool uses data from the synthetic FHIR server to train its machine learning model. The generation of synthetic data and its conversion to machine learning format is achieved via the SynthHIR system. The demonstration tool utilizes two anonymized datasets for developing the machine learning model: the Norwegian Patient Registry (NPR) \cite{npr}, which contains hospitalization details, and the Norwegian Prescription Database (NorPD) \cite{norpd}, which contains prescription details. These NPR and NorPD datasets are combined into a single CSV file containing 60,000 samples and 35 attributes. The combined dataset contains details about entities such as patient, prescriber, hospitalization, diagnosis, and prescription. More information about the dataset can be found in section 5 of the Supplementary. A generic workflow for developing any CDSS tool using the SyntHIR system is shown in Figure \ref{fig:synthir_app_development_flow}. We present the specific steps followed in our demonstration tool below, and the same steps can be easily adapted for any other tool.

\begin{enumerate}
    \item \textbf{Define mapping}: Using a dataset with any FHIR-based EHR system requires converting them to compatible formats. Therefore, the first step of the SyntHIR workflow is creating a mapping file that maps each attribute of the real dataset (CSV file) to the respective FHIR resources and their corresponding attributes.
    
    \item \textbf{Convert CSV file to FHIR resources}: Based on the mapping, the data wrangling component reads each record from the CSV file and creates a JSON-formatted list of FHIR resources using the `convert to FHIR’ API, as listed in Table \ref{tab:synthir_api_details}.
    
    \item \textbf{Upload FHIR resources}: The converted FHIR resources are uploaded to a sensitive FHIR server using the `upload’ API (as listed in Table \ref{tab:synthir_api_details}) of the FHIR adapter component. The relationship between FHIR resources for this demonstration tool is discussed in section 2 of the Supplementary.

    \item \textbf{Generate synthetic data}: In the next step, synthetic data are generated from the real datasets. The FHIR adapter component uses its `download’ API (see Table \ref{tab:synthir_api_details}) to fetch a list of FHIR resources from the sensitive FHIR server. This list is then converted into a CSV file using the `convert to CSV’ API of the Data Wrangling component. The CSV file is further sent to the Synthetic FHIR Data Generator and outputs a CSV file with 120,000 synthetic records. Note that any number of synthetic records can be generated.

    \item \textbf{Upload synthetic data}: The generated synthetic data is converted to FHIR resources using the `convert to FHIR’ API of the wrangling component. Subsequently, the FHIR resources are uploaded to the synthetic FHIR server using the adapter's `upload' API.

    \item \textbf{Train and deploy machine learning model}: A simple machine learning model is trained using the data from the synthetic FHIR server to predict the risk of hospitalization. The model accepts eight input attributes and outputs a prediction. The trained model is deployed on Microsoft Azure and can be accessed as a REST API. See section 3 of the Supplementary for details about model training. The Data Wrangling component is used as a standalone component (as shown in Figure \ref{fig:synthir_architecture}) to convert the FHIR data to a CSV file for model training. Note that the CSV file used by the model undergoes two data conversions: initially from CSV to FHIR format and subsequently back to CSV. This step could be eliminated with minor modifications in the implementation. However, we aim to introduce a generic framework that can be applied to any tool development. The synthetic data flow in SyntHIR offers two primary benefits. Firstly, it ensures that the real data is secure, as direct use of the synthetic CSV file from the generator could compromise this secure environment. Secondly, storing the generated data in an FHIR server allows data interoperability, which may not be possible if the CSV data were used directly.

    \item \textbf{Develop and deploy app}: A CDSS tool is implemented as a SMART on FHIR app and interacts with the machine learning model through the REST API.  This app is deployed on the Microsoft Azure cloud as a standalone web application. 

    \item \textbf{Connect app to synthetic FHIR server}: The app first needs to be registered as a client to access the resources of the synthetic FHIR server. The server provides a client ID and client secret to the app for authorization and authentication. Using these credentials, the app fetches the authorization code from the server and generates an access token. The app uses this token to access the FHIR resources whenever the user requests data.

    \item \textbf{Test app}: The user can enter a patient identifier on the Graphical User Interface (GUI) to get predictions from the model. Based on the identifier, the synthetic FHIR server fetches patient details. The required input attributes from the patient details are sent to the machine learning model through the REST API. The model further returns the predicted risk of hospitalization through the REST API, which is displayed to the user on the app GUI.

\end{enumerate}


\begin{figure*}[t!]
\centering
\includegraphics[width=\textwidth, height=225mm, keepaspectratio]{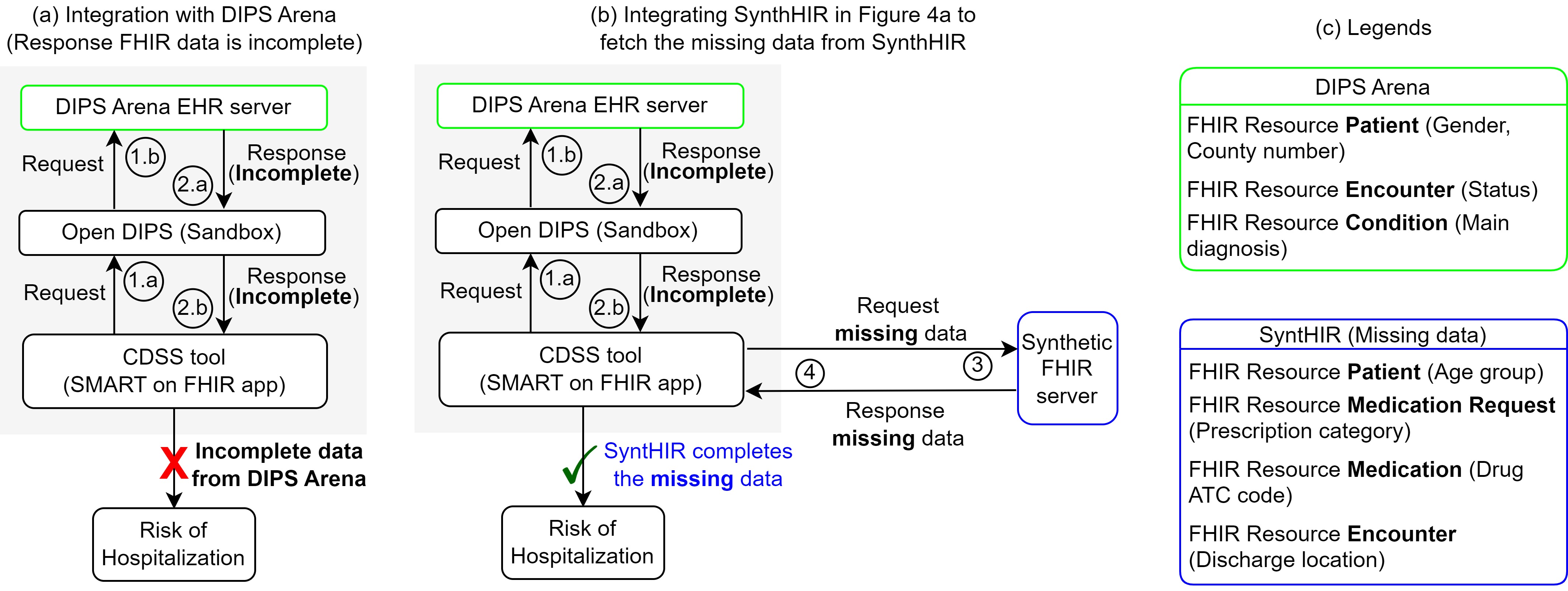}
\caption{Deploying the CDSS tool in an environment called DIPS Arena to test interoperability and transportability. \textbf{(a)} The CDSS tool is integrated with the DIPS Arena EHR server through the sandbox environment of DIPS called Open DIPS, but the response is incomplete as some missing variables are required for predicting the risk of hospitalization. \textbf{(b)} The CDSS tool is further connected to the SyntHIR synthetic FHIR server, requesting the missing variables required by the tool. \textbf{(c)} Legends denote the list of variables fetched from each DIPS Arena and SyntHIR synthetic FHIR server}
\label{fig:DIPS_Arena_deployment_steps}
\end{figure*}

\subsection{CDSS Tool Deployment in an EHR System}
\label{subsec:deploy_dips_arena}

The above demonstration tool, being interoperable (FHIR-based server) and transportable (SMART on FHIR), can be integrated into any FHIR-based EHR system. We validated this by deploying and running the tool within the open version of DIPS Arena, the latest EHR system developed by Norway’s largest EHR system vendor (DIPS). Launching the tool within the context of DIPS Arena requires registering it on the DIPS EHR server. After registration, DIPS Arena connects the tool to its EHR server by precisely following step 8 of \ref{subsec:synthir_workflow}. Additionally, DIPS configures a web page that directs to the tool's URL (tool and model remain on Microsoft Azure, as explained in steps 6 and 7 of \ref{subsec:synthir_workflow}). Users can utilize this tool as a standard application within DIPS Arena. Since DIPS supports single sign-on, users are automatically granted access without a separate login procedure. More details about the tool deployment on DIPS Arena are given in section 4 of the Supplementary.  

The tool requests data from the DIPS Arena EHR server (see Figure \ref{fig:DIPS_Arena_deployment_steps} a). Currently, DIPS Arena lacks API support for FHIR Medication profiles, namely medication and medication requests, as well as a few attributes of Patient and Encounter profiles. Therefore, the machine learning model lacks four input features required for prediction. The available and missing attributes are provided in Figure \ref{fig:DIPS_Arena_deployment_steps} c. Due to the incomplete data from DIPS Arena, the tool fails to make a prediction, as illustrated in Figure \ref{fig:DIPS_Arena_deployment_steps} a. 

The above problem can be addressed by retrieving the required missing data from the synthetic FHIR server. Therefore, we integrate the synthetic FHIR server with the tool following step 8 of \ref{subsec:synthir_workflow}. This enables the app to connect to both the DIPS Arena EHR server and the synthetic FHIR server. This ability of SyntHIR to connect the tool to multiple FHIR servers is applicable to any other tool. With access to complete data from both servers, the model successfully generates a prediction, as shown in Figure \ref{fig:DIPS_Arena_deployment_steps} b.

\begin{figure*}[t]
\centering
\includegraphics[width=\textwidth, height=225mm, keepaspectratio]{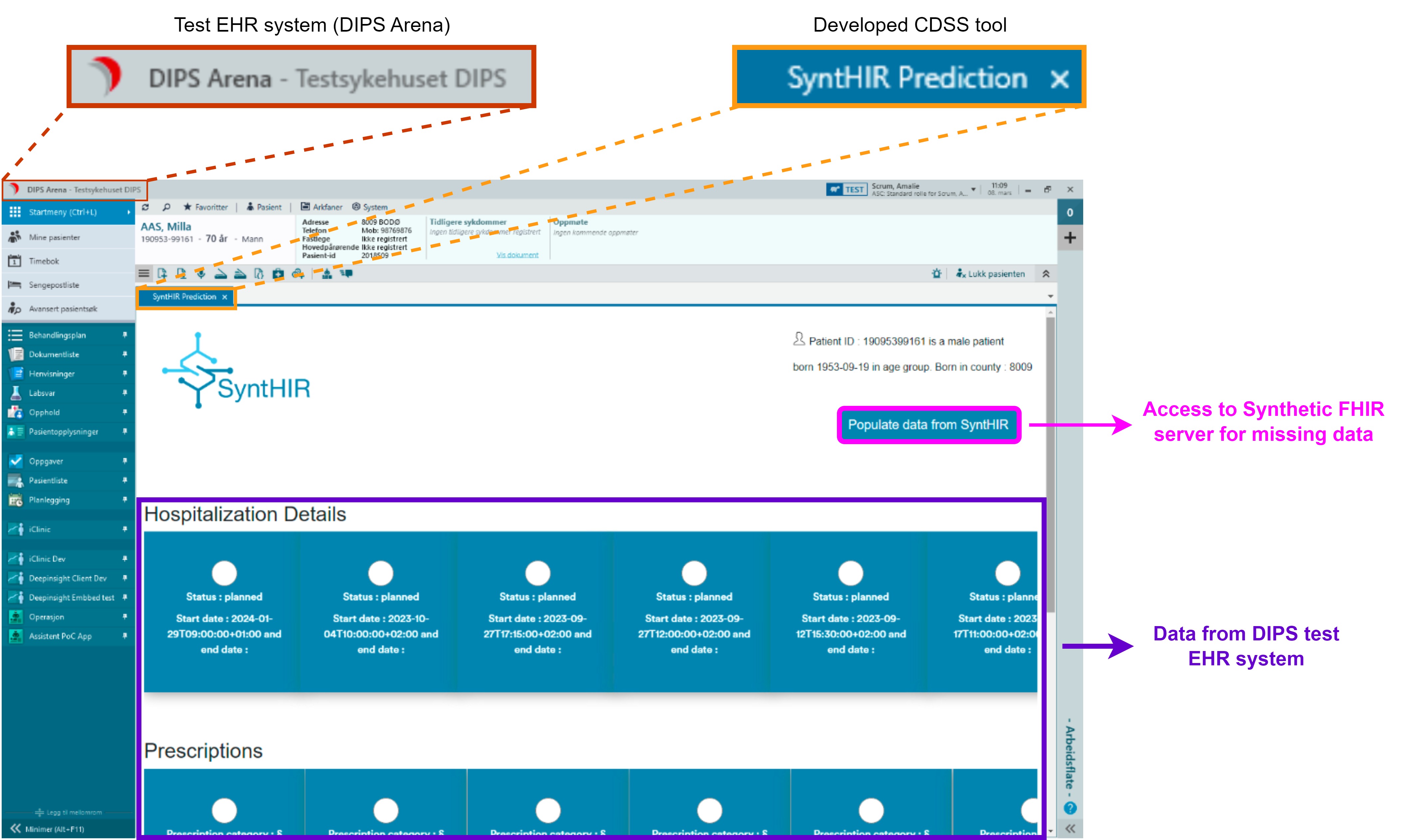}
\caption{The screenshot showcases the CDSS tool developed using SyntHIR (SyntHIR Prediction) is operational within the EHR context of DIPS Arena, the EHR test system of DIPS (`Testsykehuset'). DIPS Arena launches the SyntHIR Prediction for an individual patient, which retrieves information related to the patient profile, hospitalization records, and diagnostic codes. This information is highlighted inside the blue rectangular box in the screenshot. The missing data, including prescriptions, medications, post-hospitalization discharge location, and patient age group required by the CDSS tool, are populated from the SyntHIR synthetic FHIR server as shown by the rounded pink box in the screenshot. }
\label{fig:DIPS_Arena_deployed_CDSStool_screenshot}
\end{figure*}

The CDSS tool appears in the DIPS Arena GUI as “SyntHIR Prediction”, depicted in Figure \ref{fig:DIPS_Arena_deployed_CDSStool_screenshot}. Initially, the user launches the tool, which retrieves the patient profile, hospitalization records, and diagnostic codes from the DIPS test EHR server. As this information alone is insufficient for prediction, users can supplement it by interacting with the SyntHIR synthetic FHIR server via the “Populate data from SyntHIR” button on the GUI. 

Following authentication with the synthetic FHIR server, as outlined in step 8 of \ref{subsec:synthir_workflow}, the GUI displays a drop-down list for selecting patient age groups, discharge locations, prescriptions (categorized by prescription type), and medications (categorized by drug ATC Code). The user manually selects these fields, prompting a prediction. Although automating this selection of synthetic data is feasible, such as patient age from synthetic EHR data, we opted for manual selection for demonstration purposes.


\section{Discussion}\label{sec:discussion}

The proposed principles of the SyntHIR system address the longstanding need for a novel architecture that supports data accessibility, tool interoperability, and tool transportability of machine learning-based CDSS tools. Previous efforts by Yoo, J. et al. \cite{yoo2022development} proposed a solution by facilitating the implementation and deployment of CDSS tools in healthcare settings, specifically CANE (Common data model-based intelligent Algorithm Network Environment) settings, supporting interoperability. However, it lacks the transportability of tools to other EHR systems and does not support data accessibility. By comparison, the demonstration tool developed using SynthHIR principles can be seamlessly lifted to any FHIR-based EHR system and provide access to synthetic data. 

The tool's transportability, supported by SyntHIR, allows developers to migrate applications across FHIR-based EHR systems. Moreover, the SyntHIR workflow provides clear steps to facilitate the development, testing, and deployment of CDSS tools. The modular components of SyntHIR may further help developers in other use cases. For instance, the Data Wrangling component can be used as a standalone component outside the secure environment to convert between the machine learning format and the FHIR format.

In our study, we demonstrated SyntHIR's ability to translate a model to a CDSS tool. We focus on machine learning based CDSS tools for the SyntHIR system because they represent the most challenging aspect of decision support systems, requiring realistic data across all dimensions to ensure accurate predictions. Unlike simpler rule-based systems, which prioritize format over content and relationships, machine learning-based systems demand comprehensive data realism, making them suitable for meeting the increasing requirements of CDSS tools. Furthermore, the SyntHIR system can be extended to create other types of decision support systems, including knowledge-based \cite{gonzalez2015understanding, gholamzadeh2023application, song2021interpretable}, guideline-based \cite{goud2008development, kilsdonk2017factors}, and data-driven systems \cite{basile2023business, hayn2018predictive}.

The SyntHIR system supports testing and debugging healthcare applications based on the FHIR server, by providing synthetic data as a powerful alternative to real data for developers in the nascent stage of app development. This enables developers to grasp the nuances of real data and anticipate the app's behavior in real-world scenarios. A key strength of the SyntHIR is its ability to retrieve data from multiple sources, making it valuable for developing, testing, and validating CDSS tools in complex, multi-source environments. Synthetic data can be incorporated into automated testing pipelines to monitor the application’s performance and functionality continuously, and large volumes of such data enable stress testing to identify and resolve performance bottlenecks. 

Vendors also benefit from tool transportability, as the tool can be seamlessly deployed within their EHR system. Since the tool is deployed within the context of the EHR system, it eliminates the necessity to use external applications. This allows EHR vendors to enhance the user experience, as users can operate CDSS tools alongside other components of the EHR system. Currently, vendors offer developers a sandbox environment with limited data to develop the CDSS tool. However, by adopting the SyntHIR principles, EHR vendors can provide developers with a fully functional EHR system equipped with synthetic data, as demonstrated in the DIPS Arena experiment. 

Data owners can utilize SyntHIR to generate and disseminate a synthetic version of their real data for further research. We open-source all the components of SyntHIR \footnote{\label{githubSyntHIR}\url{https://github.com/synthir}}, where we welcome feedback from the research community to guide ongoing enhancements. This study aims to advance the translation of machine learning-based CDSS tools from research settings into clinical practice. 

We validate the SyntHIR workflow via a proof-of-concept tool. To the best of our knowledge, this workflow handles all the foreseeable user scenarios for tool development. However, some cases may require additional steps, such as validating the privacy of synthetic data and data pre-processing. Additionally, many EHR systems used by hospitals in Europe and the US have yet to adopt FHIR standards due to the lack of mandatory compliance \cite{mandl2024integration}. Consequently, the CDSS tool developed using SyntHIR may not be compatible with these EHR systems, as the Data Wrangling component exclusively supports the FHIR standard. Future work will extend the capabilities of this component to support additional data formats.

We also intend to include support for additional synthetic data generators, such as Synthea \cite{walonoski2018synthea} and Synthetic Data Vault \cite{syntheticdatavault}, and extend SyntHIR to handle unstructured data, including text and images. SyntHIR focuses on the efficient translation of the CDSS tool rather than the quality of the synthetic data. Nonetheless, assessing data quality is crucial for training machine learning models effectively. Thus, we aim to integrate a synthetic data quality evaluation module in future iterations.

\backmatter

\section*{List of abbreviations}

\begin{tabbing}
\hspace{3.2cm} \= \kill
\textbf{EHR:} \> Electronic Health Record \\
\textbf{FHIR:} \> Fast Healthcare Interoperability Resources \\
\textbf{CDSS:} \> Clinical Decision Support System \\
\textbf{SMART:} \> Substitutable Medical Applications, Reusable Technologies \\
\textbf{API:} \> Application Programming Interface \\
\textbf{URL:} \> Uniform Resource Locators \\
\textbf{CSV:} \> Comma Separated Values \\
\textbf{NPR:} \> Norwegian Patient Registry \\
\textbf{NorPD:} \> Norwegian Prescription Database
\end{tabbing}

\section*{Supplementary information}

This article is accompanied by supplementary files that include additional figures and data tables. 

\section*{Declarations}

\subsection*{Ethics approval and consent to participate}
Ethical approval for using the anonymized data to generate the synthetic dataset in our study was granted by the Regional Committees for Medical and Health Research Ethics (REK) in Norway (REK: project number 193383 and REK Sør‑Øst: 2014/2182). Informed consent from individuals was not required, as the use of anonymized data from Norwegian national health registries for research purposes is permitted by law, i.e., Health Registry Act (Helseregisterloven, Act No. 44) [https://lovdata.no/dokument/NL/lov/2008-06-20-44], and this exemption was confirmed and approved by REK. This study was conducted in accordance with the principles of the Declaration of Helsinki. 

\subsection*{Consent for publication}
Not applicable

\subsection*{Availability of data and materials}
The dataset used to demonstrate the development of the CDSS tool is composed of synthetic data generated using anonymized Norwegian Patient Registry (NPR) data, which includes hospitalization details, and the Norwegian Prescription Dataset (NorPD), which contains prescription details. We cannot share the anonymized dataset since we do not have permission from the data owner(s) to share it. This dataset is used as an example to demonstrate the development of the tool using the SyntHIR system. For the purpose of reproducibility of the results, the synthesized dataset is freely available under a public domain license at \url{https://doi.org/10.18710/YABAGM}. This license allows for the use of the data for any purpose, not limited to reproducing the results of this study.

\noindent Documentation and code for the SyntHIR system and CDSS tool developed on SMART on FHIR framework can be found at \url{https://github.com/synthir}, released under MIT license. 

\subsection*{Conflict of interest/Competing interests}
B.F. is an employee at DIPS AS, which is a company that provides EHR systems to hospitals. 

\subsection*{Funding}
This work is internally funded by the Department of Computer Science at UiT The Arctic University of Norway.

\subsection*{Author contribution}
P.C., B.F., L.A.B., and E.P. initiated the project. P.C. designed and implemented the system. K.S. provided the dataset. M.G.S.A. developed the machine learning model. P.C. and B.F. tested the tool development on DIPS Arena. P.C., B.E., E.P., L.A.B., and B.F. contributed to writing and revising the manuscript.

\bibliography{main}

\end{document}